\documentclass[10pt,twocolumn,letterpaper]{article}

\usepackage[pagenumbers]{cvpr} 

\usepackage{multirow}
\usepackage{xspace}
\usepackage{float}

\usepackage[dvipsnames]{xcolor}

\newcommand{\ours}[0]{SAM-G\xspace}
\newcommand{\oursfull}[0]{\textbf{S}egment \textbf{A}nything \textbf{M}odel for \textbf{G}eneralizable visual RL (\textbf{SAM-G})\xspace}
\newcommand{\ourwebsite}[0]{\href{https://yanjieze.com/SAM-G/}{yanjieze.com/SAM-G}\xspace}

\usepackage{colortbl}
\definecolor{ourcolor}{HTML}{06d6a0}
\definecolor{tablecolor}{HTML}{B4F6EC} 
\definecolor{tablecolor2}{HTML}{ffcdb4}
\definecolor{citecolor}{HTML}{06d6a0}

\newcommand{\down}[1]{\textcolor{gray}{\scriptsize($\downarrow #1$)}}
\newcommand{\dd}[2]{$#1\scriptstyle{\pm#2}$}
\newcommand{\ddbf}[2]{\cellcolor{tablecolor}$\mathbf{#1\scriptstyle{\pm#2}}$}

\newcommand{\ccbf}[2]{$\mathbf{#1\scriptstyle{\pm#2}}$}
\usepackage[pagebackref=true,breaklinks=true,letterpaper=true,urlcolor=citecolor,colorlinks,citecolor=citecolor,bookmarks=false,linkcolor=citecolor]{hyperref}

\definecolor{cvprblue}{rgb}{0.21,0.49,0.74}

\def\paperID{8020} 
\def\confName{CVPR}
\def\confYear{2024}

\title{Generalizable Visual Reinforcement Learning with Segment Anything Model}
\author{Ziyu Wang$^{1*}$ \quad Yanjie Ze$^{2*}$ \quad  Yifei Sun$^{23}$ \quad Zhecheng Yuan$^{124}$ \quad Huazhe Xu$^{124}$\\
$^1$Tsinghua University, IIIS \quad  $^2$Shanghai Qi Zhi Institute \quad 
$^3$Tongji University \quad 
$^4$Shanghai AI Lab  \\
$^*$Equal contribution\vspace{0.1in}\\
\ourwebsite\\
}

\begin{document}

\twocolumn[{%
\renewcommand\twocolumn[1][]{#1}%
\maketitle
\vspace{-1cm}
\begin{center}
    \centering
    \captionsetup{type=figure}
     \includegraphics[width=1.0\textwidth]{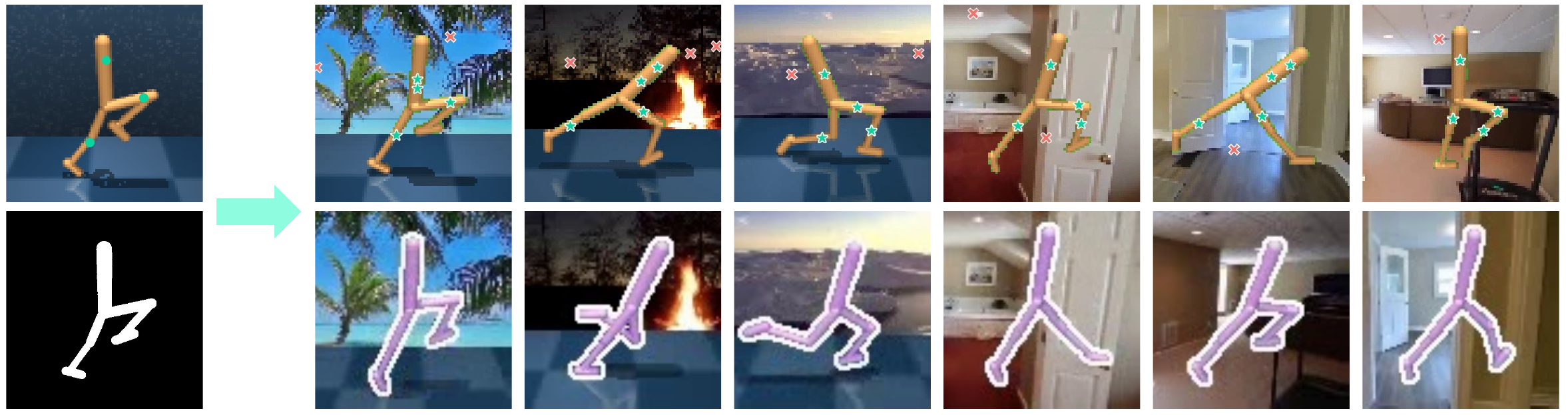}
     \vspace{-0.27in}
    \caption{\small{\textbf{Left:} For each task, one image and its mask from the training environment are provided to indicate task-relevant objects. Additionally, we assign  3 extra points to each object. \textbf{Right:} In an unseen environment, point prompts are found by correspondence, after which Segment Anything Model (SAM) produces high-quality masked images for visual RL agents.}}
    \label{fig:point prompts and masked images}
\end{center}
}]

\begin{abstract}
\vspace{-0.1in}
Learning policies that can generalize to unseen environments is a fundamental challenge in visual reinforcement learning (RL). 
While most current methods focus on acquiring robust visual representations through auxiliary supervision, pre-training, or data augmentation, the potential of modern vision foundation models remains underleveraged. 
In this work, we introduce \oursfull, a novel framework that leverages the \textit{promptable} segmentation ability of Segment Anything Model (SAM) to enhance the generalization capabilities of visual RL agents. We utilize image features from DINOv2 and SAM to find correspondence as point prompts to SAM, and then SAM produces high-quality masked images for agents directly.
Evaluated across 8 DMControl tasks and 3 Adroit tasks, \ours significantly improves the visual generalization ability without altering the RL agents' architecture but merely their observations. Notably, \ours achieves $44\%$ and $29\%$ relative improvements on the challenging \textit{video hard} setting on DMControl and Adroit respectively, compared to state-of-the-art methods. Video and code: \ourwebsite.



\end{abstract}
    
\vspace{-0.2in}
\section{Introduction}
\label{sec:intro}

Visual reinforcement learning~(RL) has achieved great success in various applications such as video games \cite{mnih2013playing,mnih2015human}, robotic manipulation~\cite{shah2021rrl,ze2023rl3d,ze2023hindex}, and robotic locomotion~\cite{yang2021learning,yang2023neural}. Despite the progress, RL agents are known to easily overfit when the training environments are not diverse and thus face severe generalization problems~\citep{cobbe2019generalization,zhang2018dissection,zhao2019investigating}.

To improve the generalization ability, recent works try to acquire a  visual representation that is robust to environment changes, by using auxiliary loss functions~\citep{hansen2021svea, bertoin2022sgqn}, data augmentation~\citep{hansen2021soda,hansen2022lfs}, and pre-training~\citep{yuan2022pieg,ze2023hindex}. 
In contrast, humans exhibit a remarkable ability to perform complex tasks in a variety of real-world scenarios, whether it is in a kitchen, a factory, or the wild, without depending on such specific designs. A plausible explanation could be our innate understanding of object concepts~\citep{baillargeon1985object}. We humans intuitively \textit{identify} and \textit{segment} task-relevant objects in cluttered environments, which is yet to be seamlessly replicated by RL-trained agents.

\begin{figure}[t]
    \centering
    \includegraphics[width=0.5\textwidth]{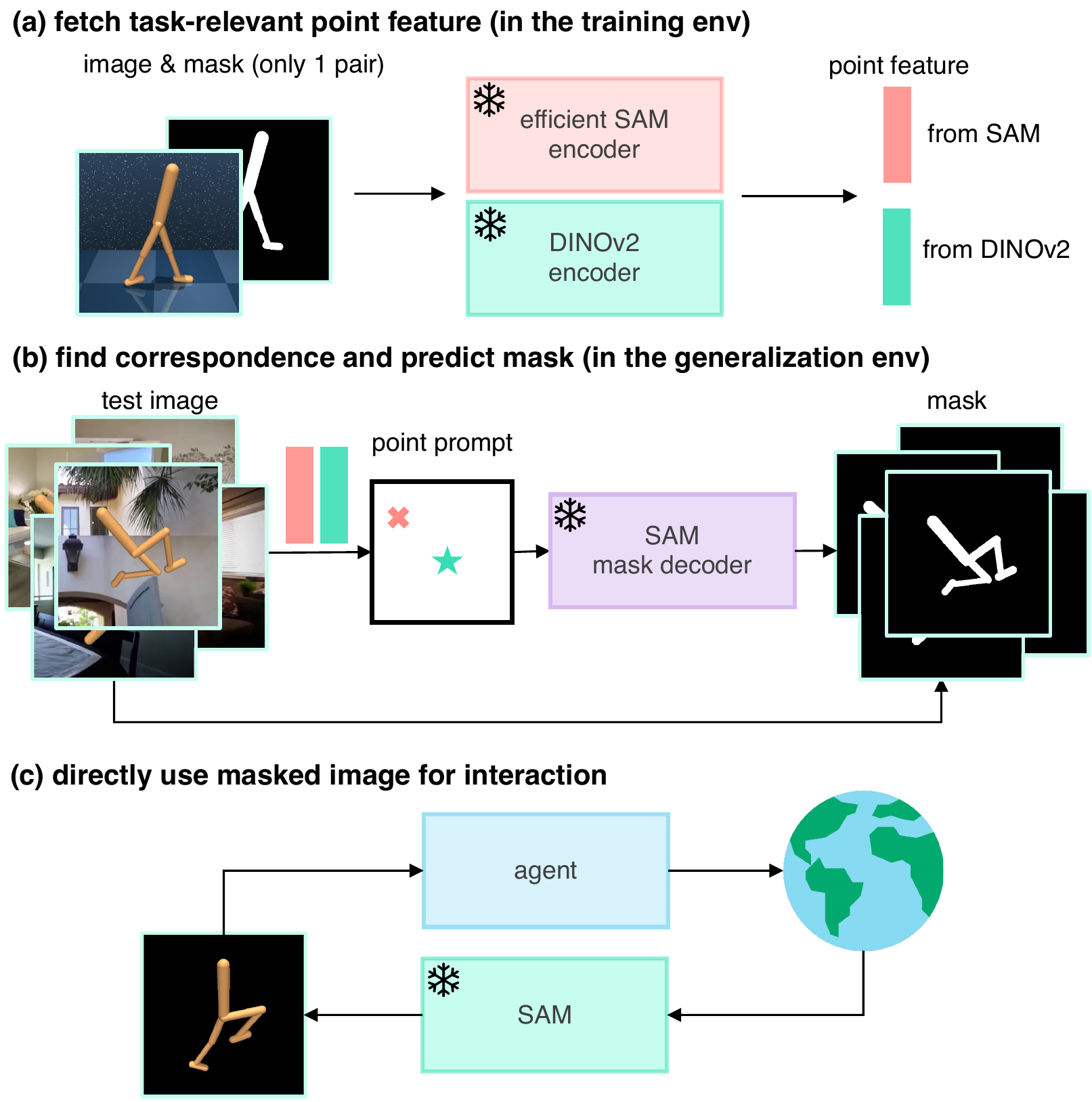}
    \vspace{-0.25in}
    \caption{\textbf{Overview of \ours.} \textbf{(a)} We provide only one image and its mask from the training environment and use encoders of vision foundation models, \textit{i.e.}, SAM~\citep{kirillov2023sam} and  DINOv2~\citep{oquab2023dinov2}, to extract point features. \textbf{(b)} We determine point prompts by finding correspondence in the test image and predict the mask with SAM. \textbf{(c)} The masked images are directly used for visual RL agents.}
    \vspace{-0.25in}
    \label{fig:overview}
\end{figure}

In this work, we equip RL agents with the capability to \textit{identify} and \textit{segment} for generalization across unseen environments, by utilizing Segment Anything Model (SAM, \citep{kirillov2023sam}), a segmentation foundation model that is promptable to receive vision prompts, such as points, bounding boxes, and languages, for user-given images. Our novel framework, \oursfull, mainly consists of two parts: \textit{identify} and \textit{segment}. We first harness image features from vision foundation models, \textit{i.e.}, DINOv2~\citep{oquab2023dinov2} and SAM, to extract task-relevant features from the training environment, termed as \textit{point feature}. Sparse points from human supervision are additionally provided to extract point features, only once for each task. Utilizing the point features, we compute the similarity map and determine point prompts in the test environment, which are then fed into SAM to accurately \textit{segment} the objects that are critical to the task at hand, with iterative mask refinement. Moreover, \ours incorporates parameter-efficient finetuning techniques~\citep{zhang2023persam} for rapid adaptation of the SAM model (only 10 seconds). Subsequently, RL agents are directly fed with these high-quality masked images in both training and generalization environments. An overview of \ours is provided in Figure~\ref{fig:overview}.

We evaluate \ours across a variety of tasks and domains, including 8 tasks from DeepMind Control Suite~\citep{tassa2018dmc} and 3 dexterous manipulation tasks from Adroit~\citep{rajeswaran2017dapg}, totaling \textbf{11} tasks. We use the generalization benchmark from DMC-GB~\citep{hansen2021soda} for DMControl tasks and RL-ViGen~\citep{yuan2023rlvigen} for Adroit tasks. Extensive experiments show that our simple yet effective framework could significantly improve visual generalization capabilities, especially for the challenging \textit{video hard} setting, where agents face dynamically changing backgrounds. Moreover, we note that \ours maintains consistent performance across settings of varying difficulty, in contrast to other baseline methods that exhibit significant performance degradation in challenging generalization settings. This observation aligns with our intuition, emphasizing the crucial importance of equipping agents with strong segmentation capabilities for robust visual generalization. 




\section{Related Work}
\label{sec:related work}

\noindent\textbf{Visual generalization in reinforcement learning.} Reinforcement learning (RL) agents are known to be facing severe generalization issues~\citep{cobbe2019generalization,zhang2018dissection,zhao2019investigating,hansen2022lfs}. Contemporary research seeks to improve the visual generalization capabilities of RL agents \citep{hansen2021soda,hansen2021svea,yuan2022pieg,ze2023rl3d, bertoin2022sgqn,yuan2023rlvigen,hansen2022lfs,yang2023movie}, with an emphasis on visual representations. \citet{hansen2021soda} propose to decouple data augmentation from the policy learning process, thereby reducing the instability that augmentation may introduce during training. \citet{hansen2021svea} stabilize Q-function learning with a two-stream architecture. \citet{yuan2022pieg} demonstrate the unexpected efficacy of utilizing an ImageNet \citep{deng2009imagenet} pre-trained encoder as representations. \citet{ze2023rl3d} pioneer the use of 3D pre-training for visual representations and facilitate a robust sim-to-real transfer. \citet{bertoin2022sgqn} employ saliency maps derived from Q-functions to guide agents in learning a mask. \citet{yang2023movie} tackle the issue of view generalization by exploiting environment dynamics.  Orthogonal to these, our work primarily focuses on harnessing the segmentation foundation model to enhance the generalization ability of RL agents. Our method could be seamlessly incorporated into other algorithms.

\noindent\textbf{Segment Anything Model} (SAM, \cite{kirillov2023sam}) is a \textit{promptable} segmentation foundation model, trained on over 11 million segmented images. It shows remarkable zero-shot segmentation abilities and can reproduce task-relevant masks with user-provided prompts, such as points, bounding boxes, and sparse masks.  While concurrent research \citep{zhang2023persam,rajivc2023sam_track,cheng2023segment_and_track} has explored the application of SAM for object localization and tracking, our work diverges in two fundamental aspects: (1) we only provide images and masks in the training environment, emphasizing generalization to any unseen environments; and (2) our objective is to enable agents to accomplish desired tasks, more than simple object tracking.

\noindent \textbf{Removing redundant information for generalization.} To achieve generalization, our intuition is to remove the redundant information across different scenes and only leave the task-relevant objects. Such intuition is shared among this work and several previous works~\citep{wang2021unsupervised_vai,fu2021learning_tia,bertoin2022sgqn,yuan2022tlda}. All these works try to achieve generalization by learning a mask: \citet{wang2021unsupervised_vai} utilize the keypoint detection and visual attention for segmentation; \citet{fu2021learning_tia} learn to reconstruct foreground and background separately; \citet{yuan2022tlda} preserve the larger Lipschitz constant areas and perform augmentation on other areas. Compared to these works, our method directly produces high-quality masks by SAM, without the need for time-consuming training, auxiliary learning objectives, or specific architecture design.

\section{Preliminaries}

\noindent\textbf{Formulation.} We model the problem as a Partially Observable Markov Decision Process (POMDP) $\mathcal{M}=\langle\mathcal{O}, \mathcal{A}, \mathcal{T}, \mathcal{R}, \gamma\rangle$, where $\mathbf{o} \in \mathcal{O}$ are high-dimensional observations (\textit{e.g.}, images), $\mathbf{a} \in \mathcal{A}$ are actions, $\mathcal{F}: \mathcal{O} \times \mathcal{A} \mapsto \mathcal{O}$ is a transition function, $r \in \mathcal{R}$ are rewards, and $\gamma \in[0,1)$ is a discount factor. During training time, the agent's goal is to learn a policy $\pi$ that maximizes discounted cumulative rewards on $\mathcal{M}$, \textit{i.e.}, $\max \mathbb{E}_{\pi}\left[\sum_{t=0}^{\infty} \gamma^t r_t\right]$. During test time, the reward signal from the environment is not accessible to agents and only observations are available. These observations are possible to experience subtle changes such as appearance changes and background changes.

\noindent\textbf{Segment Anything Model (SAM, \citep{kirillov2023sam})} is a vision foundation model designed for promptable image segmentation. Trained on  11 million labeled images, SAM has shown a notable zero-shot segmentation capability. It is particularly adept at refining predicted masks through the integration of user prompts, including sparse prompts like points, bounding boxes, and languages, and dense prompts such as masks. The \textit{promptable} nature of SAM facilitates its application in a variety of downstream tasks through prompt engineering alone, such as edge detection and object detection.

The architecture of SAM is tripartite: it features a ViT-based image encoder~\citep{dosovitskiy2020vit}, a prompt encoder, and a lightweight mask decoder. The image encoder processes images of size $1024\times 1024\times3$ into image embeddings with dimensions of $64\times 64\times256$. The prompt encoder subsequently converts user prompts into corresponding embeddings, which the mask decoder then integrates with the image embeddings to produce the segmentation \textit{mask logits}. The mask is then generated by thresholding mask logits. Inference latency is primarily attributed to the image encoder. In this work, we alleviate this issue by the usage of EfficientViT~\citep{cai2022efficientvit}.

\section{Method}

\begin{figure*}[t]
    \centering
    \includegraphics[width=1.0\textwidth]{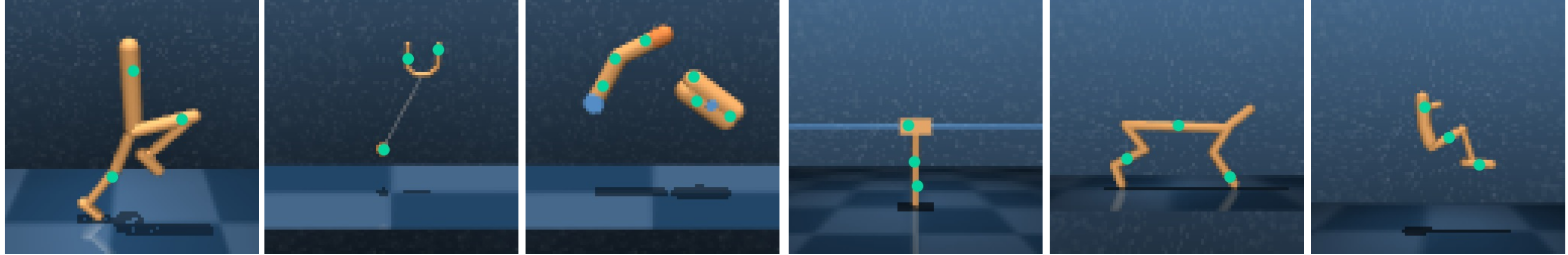}
    \vspace{-0.27in}
    \caption{\textbf{Visualization of extra points.} We use a {\color{ourcolor}green} marker to denote the extra points provided by humans. Each object is assigned with 3 extra points. It is important to note that for each task, extra points are provided only once, and the resultant point features are stored. Consequently, the time required for labeling these points can be considered negligible.}
    \label{fig:extra points}
    \vspace{-0.37cm}
\end{figure*}

\begin{figure*}[htbp]
    \centering
    \includegraphics[width=1.0\textwidth]{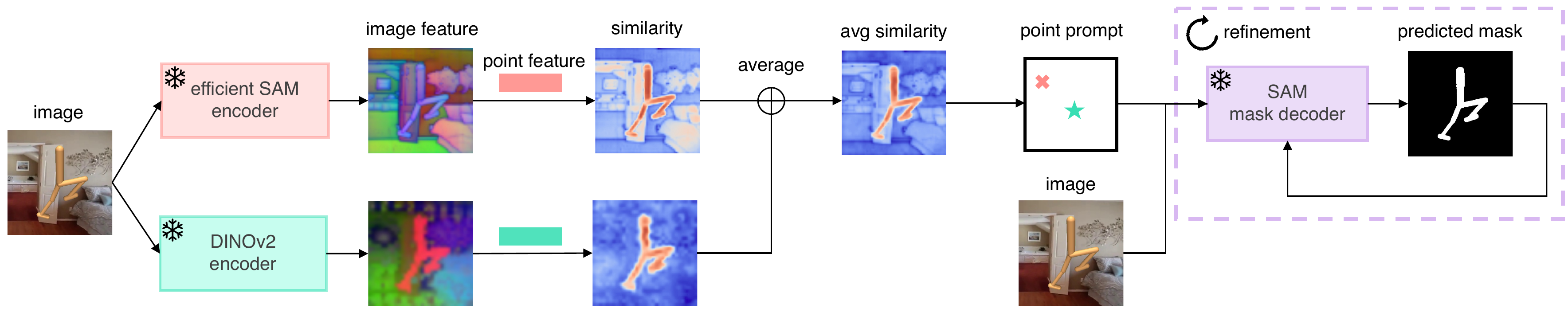}
    \vspace{-0.27in}
    \caption{\textbf{Segmentation with point feature.} We use point features from the training environment to find correspondence in the test image and obtain point prompts. Then the mask decoder iteratively refines the predicted mask given point prompts.}
    \label{fig:segmentation}
\end{figure*}

In this section, we introduce \textbf{S}egment \textbf{A}nything for Generalization (\textbf{\ours}), a framework that effectively utilizes SAM to segment out the task-relevant objects and help agents generalize to various scenes. \ours focuses on addressing the visual generalization problem of RL agents. Specifically, during the training phase, agents are exposed solely to static training environments, where the visual appearance and backgrounds are not changing. However, at test time, though the task objective does not change, the agents encounter environments with significantly altered visual properties, such as varied appearances and backgrounds. Hence, the core principle of \ours is to capture the consistent elements between the training and testing phases—namely, the object and the agent—while discounting the elements that are not pertinent to the task.

An overview of \ours is given in Figure~\ref{fig:overview}. \ours mainly consists of two parts, \textit{identify} and \textit{segment}:
\begin{itemize}
    \item \textbf{Identify.} Given only $1$ image from the training environment and its mask, we extract \textit{point feature} utilizing image features from vision foundation models. The point feature could be understood as the abstract of the task-relevant objects.
    \item \textbf{Segment.} Utilizing the point feature obtained from the training environment, we feed points found by correspondence as sparse prompts into SAM and leverage SAM to segment out task-relevant objects.
\end{itemize}
After images from environments are segmented by SAM, RL agents directly process these segmented images as input or into the replay buffer. \ours thus could be seamlessly incorporated into other visual RL algorithms.
Details of each part are illustrated in the following sections.

\subsection{Identify}
\label{section: identify}
For each task, we provide only $1$ image from the training environment and its mask as additional information to help SAM identify task-relevant objects.

\noindent\textbf{Extract image features from foundation models.} We leverage the image encoder from DINOv2~\citep{oquab2023dinov2} and SAM~\citep{kirillov2023sam} to extract image feature from the given image. Notably, for fast inference speed, we use the Efficient ViT-L1~\citep{cai2022efficientvit} architecture for SAM. We use the ViT-B/14 architecture for DINOv2.

\begin{figure}[htbp]
    \centering
    \vspace{-0.1in}
    \includegraphics[width=0.4\textwidth]{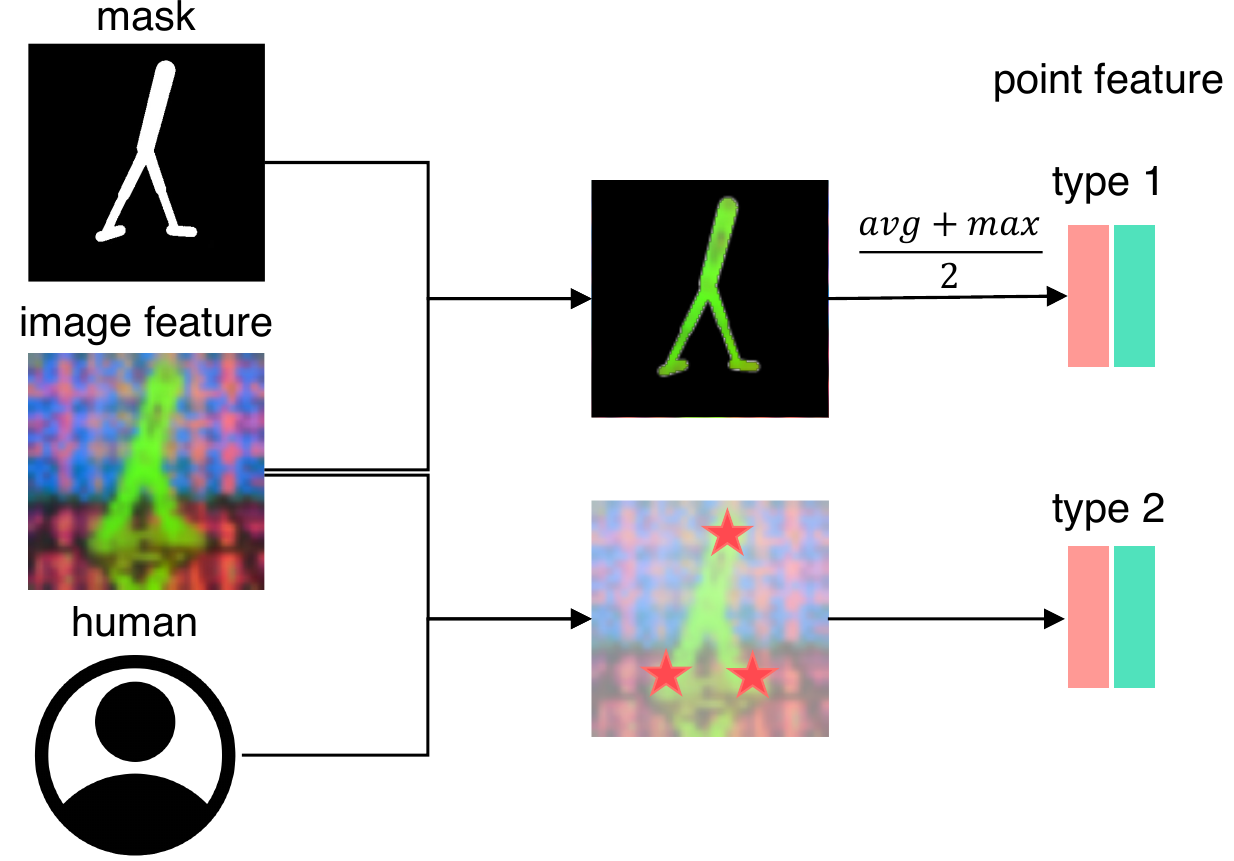}
    \vspace{-0.07in}
    \caption{\textbf{Fetch point feature.} Given only $1$ image in the training environment and its mask, we fetch two types of point features. \textit{Type 1} is computed as the average of the spatial average pooled feature and spatial max pooled feature on the masked image feature; \textit{type 2} is directly fetched given human-defined points, termed as \textit{extra points}. Each object is assigned with $3$ extra points.}
    \label{fig:fetch point feature}
    \vspace{-0.2in}
\end{figure}

\noindent\textbf{Fetch point feature.} We then fetch features that represent the task-relevant objects from image features,  termed as \textit{point feature}. Two types of point feature are extracted, as illustrated in Figure~\ref{fig:fetch point feature}: 
\begin{itemize}
    \item \textit{Type 1} point feature is automatically computed based on the masked image feature, using the average of the spatial average pooled feature and spatial max pooled feature on the masked image feature, termed as  $\frac{avg+max}{2}$.
    \item \textit{Type 2} point feature is fetched with human supervision. We manually label $3$ points for each object, termed as \textit{extra points}, visualized in Figure~\ref{fig:extra points}. Then we directly get the point feature in the corresponding coordinates. 

\end{itemize}
Take a single-object task as an example. The stored point feature has the dimension of $4\times 768$ and $4\times 256$, where $4$ means one \textit{type 1} point feature and three \textit{type 2} point features, and $768$ and $256$ are the feature dimension of DINOv2 and SAM respectively.

\begin{figure*}[t]
    \centering
    \includegraphics[width=1.0\textwidth]{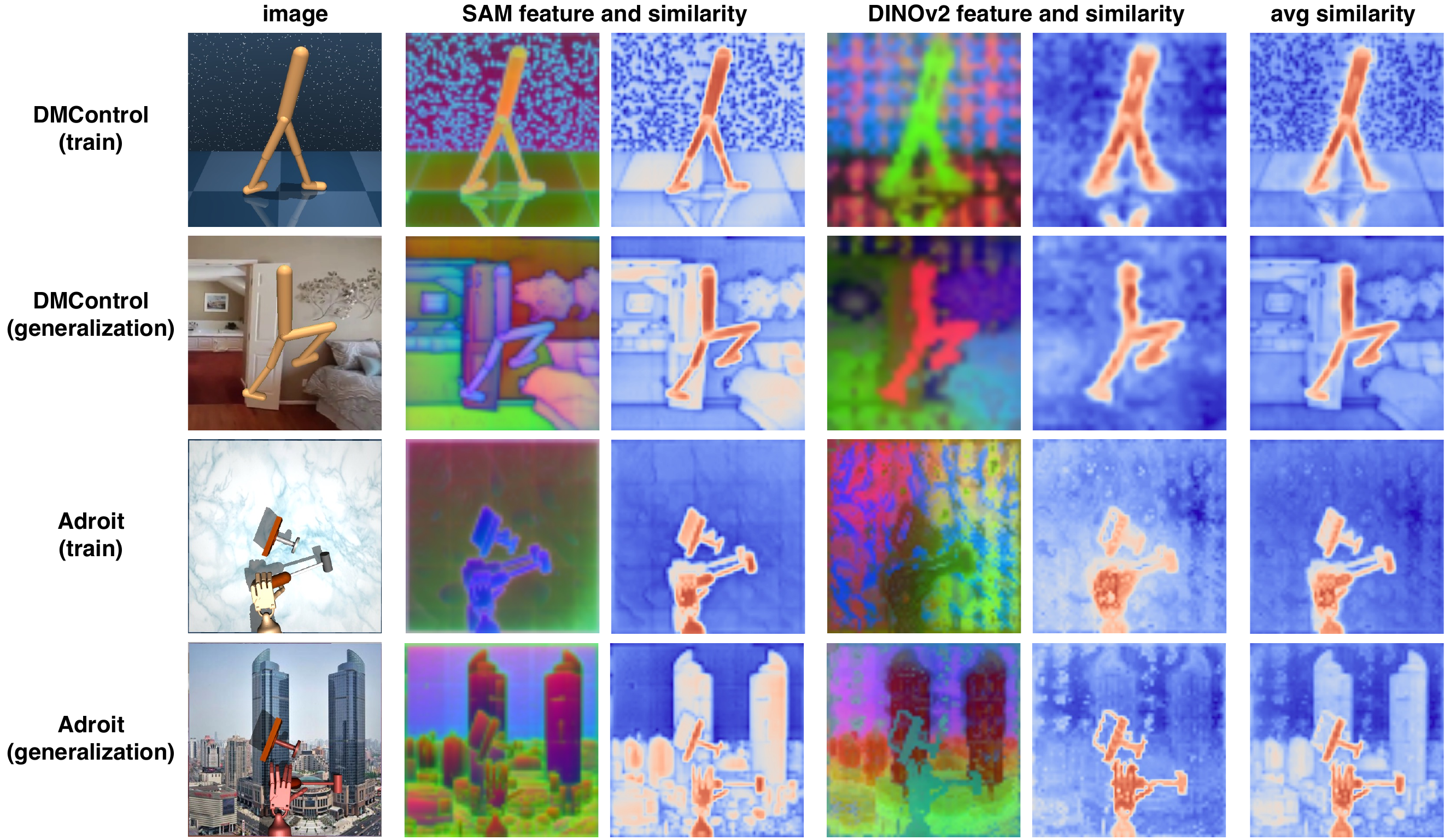}
    \vspace{-0.25in}
    \caption{\textbf{Visualization of feature maps and similarity maps.} We visualize image features from foundation models and similarity maps that are used for finding correspondence. For image features, We employ Principal Component Analysis (PCA) to reduce the dimension of image features into 3 and visualize them as RGB images. For similarity maps, areas highlighted in {\color{red}red} indicate high similarity, while {\color{blue}blue} regions represent the opposite. We only show two tasks from two domains here as the remaining tasks exhibit a similar pattern.}
    \label{fig:vis feature map}
    \vspace{-0.1in}
\end{figure*}

\subsection{Segment}

We now describe how to utilize the point feature to find point prompts and segment out the task-relevant objects with SAM. The segmentation process is the same for images from the training and generalization environments. An illustration of the segmentation process is given in Figure~\ref{fig:segmentation}.

\noindent\textbf{Determining point prompts via correspondence.} To segment a novel image, we commence by extracting its image features using the encoders of SAM and DINOv2, the same as before. Subsequently, we construct a similarity map between these newly obtained image features and the pre-established point features. For each point feature, we average the similarity maps from two foundational models to achieve better object localization. As shown in Figure~\ref{fig:segmentation}, the averaged similarity map distinctly highlights the target object. For the similarity map corresponding to the \textit{type 1} point feature, we identify the point of highest similarity to serve as a positive point prompt and the point of least similarity as a negative point. In contrast, for the similarity map constructed by the \textit{type 2} point feature, we solely select the point of highest similarity as a positive point prompt. This is because  \textit{type 2} point features only represent information from a specific part of the object, and the point of least similarity could be mistakenly put on other parts of the object.  Features and similarity maps are visualized in Figure~\ref{fig:vis feature map}.

\noindent\textbf{Mask prediction and refinement.} After point prompts are obtained and encoded into embeddings by the prompt encoder of SAM, the mask decoder decodes the image feature and the prompt embedding into mask logits~\citep{kirillov2023sam}. We further refine the prediction by repeating the mask decoding process with additional prompts. Specifically, after the 1st mask prediction, we compute the bounding box of the mask and select a point of least similarity in this box, based on the similarity map we obtain previously. Then we perform the 2nd and 3rd mask prediction with old prompts and newly added point prompts. Note that iterative mask refinement has been used in previous works such as SAM~\citep{kirillov2023sam}, PerSAM~\citep{zhang2023persam}, and SAM-PT~\citep{rajivc2023sam_track}, while we propose to add additional negative points based on the similarity map in each refinement, to better separate the foreground and background in the generalization setting. The point prompts and masked images are visualized in Figure~\ref{fig:point prompts and masked images}.


\noindent\textbf{One shot adaptation.} Before RL agents start to loop in the training environments, we fast adapt only two weights in SAM with our initially provided image and mask, following PerSAM~\citep{zhang2023persam}. Specifically, SAM produces $3$ mask logits, denoted as $M_1,M_2,M_3$, and conducts a weighted summation,
$$
M=w_1 \cdot M_1+w_2 \cdot M_2+\left(1-w_1-w_2\right) \cdot M_3\,.
$$
We use the provided image from the training environment as input and its mask as the supervision to adjust the weights $w_1$ and $w_2$. All other parts of SAM are frozen. This process takes roughly $10$ seconds on an Nvidia 3090 GPU, thus its time consumption could be almost neglected.

\section{Experiments}
In this section, we evaluate the generalization ability of \ours on $11$ tasks, including $8$ tasks from DMControl~\citep{tassa2018dmc} and $3$ dexterous manipulation tasks from Adroit~\citep{rajeswaran2017dapg}. We use the generalization setting from DMControl Generalization Benchmark (DMC-GB, \citep{hansen2021soda}), where four settings are considered with increasing difficulty: \textit{color easy}, \textit{color hard}, \textit{video easy}, and \textit{video hard}. Detailed descriptions of tasks and generalization settings are in Appendix~\ref{appendix: task and generalization description}.  \textbf{Videos are given in supplementary files.}

\begin{figure*}[t]
    \centering
    \includegraphics[width=1.0\textwidth]{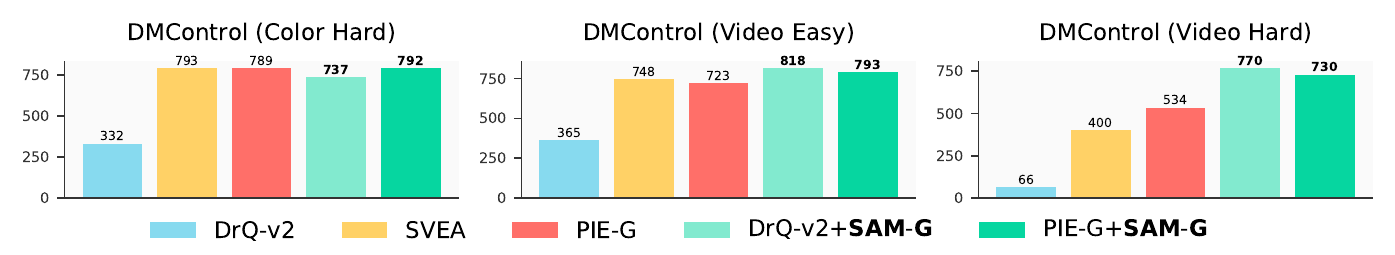}
     \includegraphics[width=1.0\textwidth]{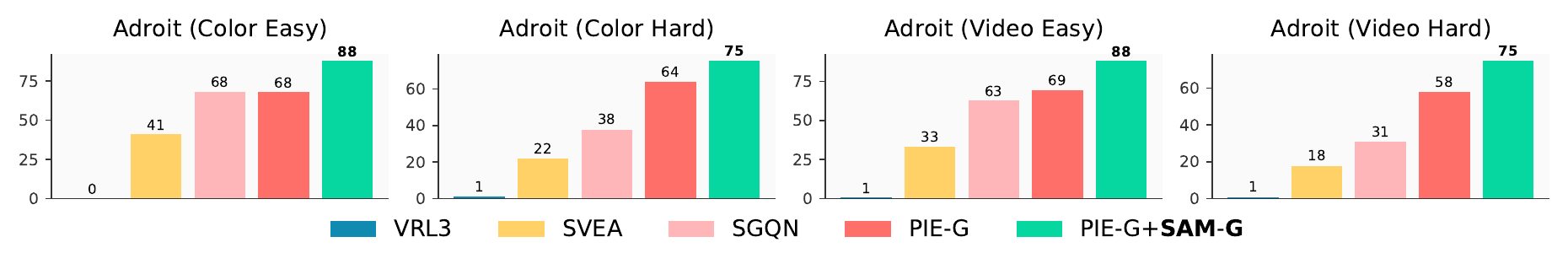}
     \vspace{-0.3in}
    \caption{\textbf{Visual generalization results across 2 domains and 4 settings.} Our method \ours could robustly improve the visual generalization ability of visual RL agents such as DrQ-v2~\citep{yarats2021drqv2} and PIE-G~\citep{yuan2022pieg}. Notably, in the challenging \textit{video hard} setting, \ours surpasses previous state-of-the-art method PIE-G with $44\%$ and $29\%$ relative improvement on DMControl and Adroit respectively.}
    \label{fig:experiment summary}
\end{figure*}

\subsection{Experiment Setup}

\noindent\textbf{Baselines.} We benchmark \ours against the following strong baselines: (1) \textbf{DrQ-v2}~\citep{yarats2021drqv2} that applies random shift as data augmentation; (2) \textbf{VRL3}~\citep{wang2022vrl3} that uses offline RL to pre-train for sample efficiency; (3) \textbf{SVEA}~\citep{hansen2021svea} that stabilizes the Q-function learning via an auxiliary loss; (4) \textbf{SGQN}~\citep{bertoin2022sgqn} that leverages the saliency map to help remove redundant information; (5) \textbf{PIE-G}~\citep{yuan2022pieg} that applies a pre-trained image encoder as the representation. Among these baselines, DrQ-v2 and VRL3 are designed for sample efficiency and thus their generalization ability is limited; DrQ-v2 and VRL3 serve as the backbone algorithms on DMControl tasks and VRL3 respectively. Other algorithms, including SVEA, PIE-G, SGQN, and \ours, are designed for visual generalization and built upon these backbone algorithms.

\noindent\textbf{Implementation.} We implement \ours on top of DrQ-v2,  PIE-G (DrQ-v2 version), and PIE-G (VRL3 version), to show that \ours could be incorporated into other visual RL agents seamlessly. Visual observations are a stack of 3 RGB frames of size $84\times 84 \times 3$. Mean and standard deviation of $3$ random seeds are reported. For Adroit tasks, we use the implementation from RL-ViGen~\citep{yuan2023rlvigen} since DMC-GB does not provide generalization settings on Adroit tasks. Hyperparameters are given in the supplementary material.

\subsection{Main Experiments}
Considering the extensive scale of our experiments, we present a summary of our main generalization results in Figure~\ref{fig:experiment summary}, across 2 domains and 4 settings. We then detail our findings below.

\begin{figure*}[t]
    \centering
    \includegraphics[width=1.0\textwidth]{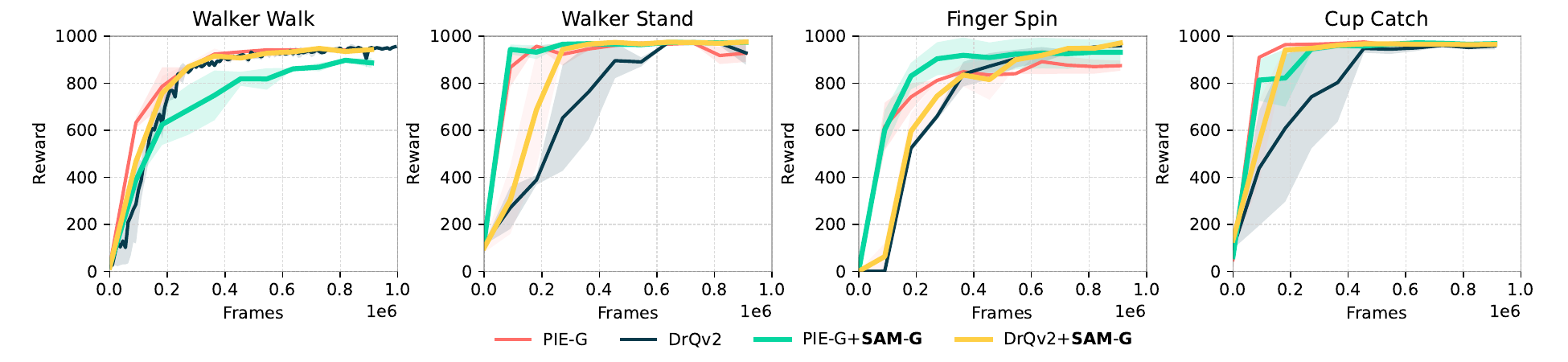}
    \includegraphics[width=1.0\textwidth]{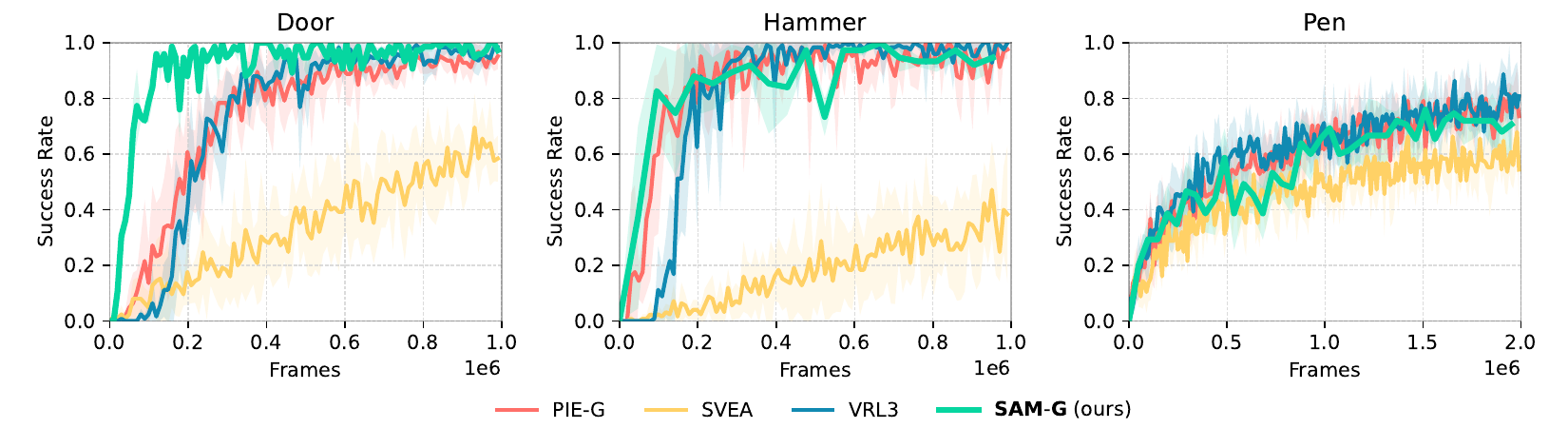}
    \vspace{-0.3in}
    \caption{\textbf{Training curves on Adroit and DMControl.} Mean of 3 random seeds, shaded area is $\pm1$ std. We observe that \ours (PIE-G version) achieves better sample efficiency on Door and competitive results on Hammer and Pen compared to other strong baselines. On DMControl, \ours (both DrQ-v2 version and PIE-G version) achieves compared efficiency to other strong baselines.This indicates that applying masked images directly would not negatively impact the training performance.}
    \label{fig:training curves}
    \vspace{-0.1in}
\end{figure*}

\noindent\textbf{DMControl.}  Detailed generalization results on 8 DMControl tasks are given in Table~\ref{table:dmc}. We omit the \textit{color easy} setting since this setting is too easy for all methods on DMControl. Training curves of 4 tasks from DMControl are displayed in Figure~\ref{fig:training curves}, where all the algorithms achieve similar sample efficiency and convergence. We could also observe that \ours slightly reduces the variance during training. Our observations are summarized as follows:


\begin{itemize}
    \item Without specific design for generalization, the backbone algorithm DrQ-v2 encounters difficulties in all generalization settings. Other generalization methods such as SVEA and PIE-G exhibit robust performance in relatively easy scenarios, such as \textit{color hard} and \textit{video easy}, but exhibit significant performance drops in the more challenging \textit{video hard} setting. This shows the limitations of current visual reinforcement learning algorithms and underscores the critical role of visual generalization.
    \item In contrast, \ours consistently delivers robust performance across all settings. Notably, in the \textit{video hard} setting where other methods struggle, \ours achieves $770$ average scores, largely surpassing PIE-G with $44\%$ relative improvements. This verifies our intuition that the robust generalization capacity could arise from the robust segmentation capacity. Leveraging the capabilities of strong segmentation models such as SAM, agents should be adept at handling challenging generalization scenarios.
\end{itemize}

\begin{table}[t]
\centering
\caption{\textbf{Generalization results on DMControl.} Mean and std of 3 seeds are reported. Generalization settings are from DMC-GB~\citep{hansen2021soda}.}
\vspace{-0.1in}
\label{table:dmc}
\resizebox{0.47\textwidth}{!}{%
\begin{tabular}{lcccccc}
\toprule

DMControl  &  \multirow{2}{*}{DrQ-v2} & \multirow{2}{*}{SVEA} & \multirow{2}{*}{PIE-G} & \textbf{DrQ-v2} & \textbf{PIE-G}\\
 (color hard)& & & & \textbf{+ \ours}  & \textbf{+ \ours} &\\
\midrule
Walker Walk & \dd{168}{90} & \dd{760}{145} & \dd{884}{20} & \dd{805}{37} & \ddbf{895}{24}\\
Walker Stand & \dd{413}{61} & \dd{942}{26} & \dd{960}{15} & \dd{839}{80} & \ddbf{971}{4} \\
Cartpole Swingup & \dd{277}{80} & \ccbf{837}{23} & \dd{749}{46} & \dd{728}{15} & \dd{751}{57}\\
Cheetah Run & \dd{109}{45} & \dd{273}{23} & \ccbf{369}{53} & \dd{280}{48} & \dd{349}{28}\\
Hopper Stand & \dd{383}{41} & \ccbf{715}{78} & \dd{681}{72} &\dd{659}{7}& \dd{706}{55}\\
Finger Spin & \dd{676}{134} & \dd{977}{5} & \dd{882}{69} & \dd{740}{32} & \ddbf{935}{66}\\
Ball\_in\_cup Catch & \dd{469}{99} & \dd{961}{7} & \dd{964}{7} & \dd{949}{26} & \ddbf{970}{5} \\
Quadruped Walk & \dd{162}{40} & \dd{879}{8} & \dd{822}{8} & \ddbf{893}{19} & \dd{759}{44} \\
\midrule
\textbf{Average} & $332$ & $\mathbf{793}$ & $789$ & $737$ & $792$ \\
\bottomrule
\end{tabular}}

\vspace{0.05in}
\resizebox{0.47\textwidth}{!}{%
\begin{tabular}{lcccccc}
\toprule

DMControl  &  \multirow{2}{*}{DrQ-v2} & \multirow{2}{*}{SVEA} & \multirow{2}{*}{PIE-G} & \textbf{DrQ-v2} & \textbf{PIE-G}\\
 (video easy)& & & & \textbf{+ \ours}  & \textbf{+ \ours} &\\
\midrule
Walker Walk & \dd{175}{117} & \dd{819}{71} & \dd{871}{22} & \dd{887}{33} & \ddbf{930}{28}\\
Walker Stand & \dd{560}{48} & \dd{961}{8} & \dd{957}{12} &\dd{839}{80} & \ddbf{970}{7} \\
Cartpole Swingup & \dd{267}{41} & \dd{782}{27} & \dd{587}{61} & \ddbf{855}{7} & \dd{710}{90}\\
Cheetah Run & \dd{64}{22} & \dd{249}{20} & \dd{287}{20} & \dd{282}{49} & \ddbf{345}{8}\\
Hopper Stand & \dd{261}{62} & \dd{678}{30} & \dd{514}{51} &\ddbf{971}{2} & \dd{715}{71} \\
Finger Spin & \dd{456}{15} & \dd{808}{33} & \dd{837}{107} & \ddbf{937}{21} & \dd{928}{65} \\
Ball\_in\_cup Catch & \dd{454}{60} & \dd{871}{106} & \dd{922}{20} & \dd{858}{79} & \ddbf{957}{8} \\
Quadruped Walk & \dd{681}{29} & \dd{818}{6} & \dd{805}{14} & \ddbf{916}{14} & \dd{789}{72} \\
\midrule
\textbf{Average} & $365$ & $748$ & $723$ & $\mathbf{818}$ & $793$\\
\bottomrule
\end{tabular}}

\vspace{0.05in}
\resizebox{0.47\textwidth}{!}{%
\begin{tabular}{lcccccc}
\toprule

DMControl  &  \multirow{2}{*}{DrQ-v2} & \multirow{2}{*}{SVEA} & \multirow{2}{*}{PIE-G} & \textbf{DrQ-v2} & \textbf{PIE-G}\\
 (video hard)& & & & \textbf{+ \ours}  & \textbf{+ \ours} &\\
\midrule
Walker Walk & \dd{34}{11} & \dd{377}{93} & \dd{600}{28} & \dd{846}{38} & \ddbf{857}{31}\\
Walker Stand & \dd{151}{13} & \dd{834}{46} & \dd{852}{56} &\dd{966}{2} & \ddbf{964}{4} \\
Cartpole Swingup & \dd{130}{3} & \dd{393}{45} & \dd{401}{21} & \ddbf{747}{5} & \dd{653}{58}\\
Cheetah Run & \dd{23}{5} & \dd{105}{37} & \dd{154}{17} & \dd{283}{46} & \ddbf{327}{11}\\
Hopper Stand &\dd{5}{5} &\dd{221}{13} & \dd{235}{17}&\ddbf{840}{30} & \dd{618}{63} \\
Finger Spin & \dd{21}{4} & \dd{335}{58} & \dd{762}{59} & \ddbf{932}{22} & \dd{908}{66} \\
Ball\_in\_cup Catch & \dd{97}{27} & \dd{403}{174} & \dd{786}{47} & \dd{771}{101} & \ddbf{880}{54}\\
Quadruped Walk & \dd{69}{21} & \dd{532}{23} & \dd{483}{62} & \ddbf{774}{22} & \dd{635}{21} \\
\midrule
\textbf{Average} & $66$ & $400$ & $534$ & $\mathbf{770}$ & $730$\\
\bottomrule
\end{tabular}}
\vspace{-0.1in}
\end{table}

\noindent\textbf{Adroit.} We present the generalization results for 3 Adroit tasks in Table~\ref{table:adroit}, spanning four challenging settings of increasing difficulty. The training curves for the Adroit tasks are displayed in Figure~\ref{fig:training curves}, where all algorithms exhibit similar convergence patterns, except for SVEA. This ensures a fair comparison in the generalization benchmark. Notably, we observe that \ours significantly improves sample efficiency on the Door task and does not hinder learning on the other two tasks. Our observations regarding the generalization results in Table~\ref{table:adroit} are summarized as follows:
\begin{itemize}
    \item \ours demonstrates substantial performance enhancements, notably surpassing the previous state-of-the-art algorithm, PIE-G. This result is particularly significant because each Adroit task involves multiple objects, highlighting the generality of \ours for multi-object tasks, extending beyond the simpler single-object tasks in DMControl.
    \item The performance of other baselines, including the previous state-of-the-art algorithm SVEA, is subpar. This once again underscores the critical importance of robust visual generalization.
\end{itemize}

\begin{table}[t]
\centering
\caption{\textbf{Generalization results on Adroit.} Mean and std of 3 seeds are reported. Generalization settings are from RL-ViGen~\citep{yuan2023rlvigen}. }
\vspace{-0.1in}
\label{table:adroit}

\resizebox{0.47\textwidth}{!}{%
\begin{tabular}{lcccccc}
\toprule
Adroit  &  \multirow{2}{*}{VRL3} & \multirow{2}{*}{SVEA} & \multirow{2}{*}{SGQN} & \multirow{2}{*}{PIE-G} & \textbf{PIE-G}\\
 (color easy)& & & &  & \textbf{+ \ours} &\\
\midrule
Pen & \dd{1.7}{0.7} & \dd{53.3}{7.6} & \dd{71.3}{5.0} & \dd{76.0}{7.0} & 
\ddbf{79.0}{4.4}\\
Door & \dd{0.0}{0.0} & \dd{45.4}{9.7} & \dd{58.2}{12.9} & \dd{81.6}{6.7} & \ddbf{96.0}{3.6}\\
Hammer & \dd{0.0}{0.0}  & \dd{24.0}{15.6} & \dd{75.0}{8.6}& \dd{45.8}{19.3} & \ddbf{88.0}{9.3}\\

\midrule
\textbf{Average} & $0.6$ & $40.9$ & $68.2$ & $67.8$ & $\mathbf{87.7}$ \\
\bottomrule
\end{tabular}}

\vspace{0.05in}

\resizebox{0.47\textwidth}{!}{%
\begin{tabular}{lcccccc}
\toprule

Adroit  &  \multirow{2}{*}{VRL3} & \multirow{2}{*}{SVEA} & \multirow{2}{*}{SGQN} & \multirow{2}{*}{PIE-G} & \textbf{PIE-G}\\
 (color hard)& & & &  & \textbf{+ \ours} &\\
\midrule
 
Pen & \dd{3.7}{2.1} & \dd{44.7}{5.8} & \dd{54.3}{7.1} & \dd{70.3}{4.6} & 
\ddbf{75.3}{6.7}\\
Door & \dd{0.0}{0.0} & \dd{11.8}{3.6} & \dd{31.4}{9.8} & \dd{67.8}{9.8} & \ddbf{90.3}{4.2}\\
Hammer & \dd{0.0}{0.0} & \dd{9.0}{6.3} & \dd{27.8}{6.4} & \dd{53.2}{12.5} & \ddbf{60.0}{12.8}\\
\midrule
\textbf{Average} & $1.2$& $21.8$& $37.8$& $63.8$& $\mathbf{75.2}$\\

\bottomrule
\end{tabular}}

\vspace{0.05in}

\resizebox{0.47\textwidth}{!}{%
\begin{tabular}{lcccccc}
\toprule

Adroit  &  \multirow{2}{*}{VRL3} & \multirow{2}{*}{SVEA} & \multirow{2}{*}{SGQN} & \multirow{2}{*}{PIE-G} & \textbf{PIE-G}\\
 (video easy)& & & &  & \textbf{+ \ours} &\\
\midrule
 
Pen & \dd{1.7}{0.6} & \dd{46.7}{3.8} & \dd{68.7}{8.1} & \dd{76.0}{1.7} & 
\ddbf{82.3}{5.1} \\
Door & \dd{0.0}{0.0} & \dd{44.8}{8.5} & \dd{58.2}{12.3} & \dd{81.6}{4.4} & \ddbf{98.0}{1.0} \\
Hammer & \dd{0.0}{0.0}  & \dd{8.4}{8.6} & \dd{61.0}{9.4} & \dd{50.4}{21.2} & \ddbf{83.3}{7.0} \\
\midrule
\textbf{Average} & $0.6$& $33.3$& $62.6$& $69.3$& $\mathbf{87.9}$\\

\bottomrule
\end{tabular}}

\vspace{0.05in}

\resizebox{0.47\textwidth}{!}{%
\begin{tabular}{lcccccc}
\toprule

Adroit  &  \multirow{2}{*}{VRL3} & \multirow{2}{*}{SVEA} & \multirow{2}{*}{SGQN} & \multirow{2}{*}{PIE-G} & \textbf{PIE-G}\\
 (video hard)& & & &  & \textbf{+ \ours} &\\
\midrule
 
Pen & \dd{2.7}{1.5} & \dd{41.7}{6.1} & \dd{52.3}{0.6} & \dd{60.7}{6.0} & 
\ddbf{76.7}{2.1} \\
Door & \dd{0.0}{0.0} & \dd{7.6}{1.8} & \dd{21.6}{6.4} & \dd{60.4}{12.3} & \ddbf{88.3}{3.1} \\
Hammer & \dd{0.0}{0.0} & \dd{4.2}{3.7} & \dd{19.2}{7.4} & \dd{52.6}{10.2} & \ddbf{58.7}{11.2} \\
\midrule
\textbf{Average}& $0.9$& $17.8$& $31.0$& $57.9$& $\mathbf{74.6}$\\

\bottomrule
\end{tabular}}
\end{table}

\subsection{Imitation Learning}
Since \ours could be viewed as a general framework for generalizable visuomotor policy learning, we conduct initial experiments that apply \ours for imitation learning (IL) on challenging Adroit tasks. 

\noindent\textbf{Setup.} We use the policy architecture of PIE-G~\citep{yuan2022pieg}. We collect $300$ expert demonstrations for each task using RL agents and train $80$ epochs to ensure convergence.  The training objective is simply behavior cloning with a mean squared error. We use Adam~\citep{kingma2014adam} optimizer with the learning rate $5\times 10^{-5}$ and the batch size $256$. The data augmentation in PIE-G is also applied. Our baseline is the policy part of PIE-G, which is called PIE-G as well in this section.

\noindent\textbf{Results.} As indicated in Table~\ref{table:imitation learning on adroit}, our observations on Adroit tasks remain consistent across both RL and IL settings. In both cases, \ours outperforms PIE-G by a significant margin on Door and Hammer tasks and achieves results that are comparable to PIE-G on the Pen task. These preliminary experiments in IL highlight the potential of our framework for visuomotor policy learning.

\begin{table}[t]
\centering
\caption{\textbf{Imitation learning and generalization results on Adroit.} Mean and std of 3 seeds are reported. Generalization settings are the same as the RL experiments.}
\vspace{-0.1in}
\label{table:imitation learning on adroit}
\resizebox{0.5\textwidth}{!}{%
\begin{tabular}{lccccc|c}
\toprule

Imitation &  \multirow{2}{*}{Train} & Color & Color  & Video & Video & \multirow{2}{*}{\textbf{Average}}\\

(Door) & & easy & hard & easy & hard \\
\midrule
PIE-G & \dd{93.0}{3.6}  & \dd{85.7}{2.3} &\dd{62.5}{16.5} & \dd{78.7}{7.5} & \dd{61.0}{8.5} & $76.2$ \\

\textbf{\ours} & \dd{96.7}{3.5} & \dd{97.0}{3.0} & \dd{91.3}{8.1} &  \dd{96.3}{0.6} & \dd{88.3}{1.5} & $\mathbf{93.9}$  \\
\bottomrule
\end{tabular}}

\vspace{0.05in}

\resizebox{0.5\textwidth}{!}{%
\begin{tabular}{lccccc|c}
\toprule

Imitation &  \multirow{2}{*}{Train} & Color & Color  & Video & Video & \multirow{2}{*}{\textbf{Average}}\\

(Hammer) & & easy & hard & easy & hard \\
\midrule
PIE-G & \dd{96.7}{4.2} & \dd{75.3}{4.2} & \dd{44.3}{4.0} & \dd{55.0}{6.2} & \dd{48.0}{5.3} & $63.9$\\

\textbf{\ours} & \dd{96.7}{3.5} & \dd{75.0}{2.0} & \dd{63.7}{3.5} & \dd{62.0}{4.4} & \dd{55.0}{4.4} & $\mathbf{70.5}$\\
\bottomrule
\end{tabular}}

\vspace{0.05in}

\resizebox{0.5\textwidth}{!}{%
\begin{tabular}{lccccc|c}
\toprule

Imitation &  \multirow{2}{*}{Train} & Color & Color  & Video & Video & \multirow{2}{*}{\textbf{Average}}\\

(Pen) & & easy & hard & easy & hard \\
\midrule
PIE-G & \dd{76.7}{4.2} & \dd{62.7}{6.1} & \dd{57.3}{9.5} & \dd{61.3}{3.2} & \dd{50.7}{1.2} & $61.7$\\

\textbf{\ours} & \dd{72.7}{1.2} & \dd{65.3}{5.0} & \dd{57.0}{7.0} & \dd{62.7}{11.0} & \dd{57.3}{9.0} & $\mathbf{63.0}$\\
\bottomrule
\end{tabular}}
\end{table}

\subsection{Ablations}

\begin{table}[t]
\centering
\caption{\textbf{Ablations on DMControl.} Three main design choices of \ours are ablated: using DINOv2 to extract image features; adding extra points for more point features; and iterative mask refinement. We remove each component from \ours and observe performance drops compared to full \ours.}
\vspace{-0.1in}
\label{table:ablations}
\resizebox{0.47\textwidth}{!}{%
\begin{tabular}{lcccc|l}
\toprule

DMControl & Walker  & Cartpole  & Finger & Ball\_in\_cup  & \multirow{2}{*}{\textbf{Average}}\\
(color hard) &Walk & Swingup & Spin & Catch \\
 
\midrule
\textbf{\ours} & \dd{805}{37}& \dd{728}{51}& \dd{739}{32}& \dd{949}{26} & $805$ 
\\w/o. DINOv2 &   \dd{726}{39}&  \dd{589}{75} & \dd{785}{90}& \dd{921}{10} & $755$ \down{50}
\\w/o. extra points &  \dd{761}{64} & \dd{649}{78} & \dd{783}{34}& \dd{929}{17} & $781$  \down{24}
\\w/o. refinement & \dd{708}{73} & \dd{790}{123}& \dd{617}{27} & \dd{935}{15} & $763$ \down{42}  
\\w/o. PerSam adapt & \dd{659}{93} & \dd{501}{28} & \dd{833}{136} & \dd{894}{91} & $722$ \down{172} \\

\bottomrule
\end{tabular}}

\vspace{0.05in}

\resizebox{0.47\textwidth}{!}{%
\begin{tabular}{lcccc|l}
\toprule

DMControl & Walker  & Cartpole  & Finger & Ball\_in\_cup  & \multirow{2}{*}{\textbf{Average}}\\
(video easy) &Walk & Swingup & Spin & Catch \\
 
\midrule
\textbf{\ours} & \dd{887}{33}& \dd{855}{6}& \dd{937}{21}& \dd{858}{79} & $884$
\\w/o. DINOv2 &   \dd{871}{13}&  \dd{611}{82} & \dd{949}{15}& \dd{706}{59} & $784$ \down{100}
\\w/o. extra points &  \dd{843}{55} & \dd {659}{58} & \dd{930}{24}& \dd {885}{18} & $829$  \down{55}
\\w/o. refinement & \dd{834}{136} & \dd{791}{39}& \dd{680}{24}& \dd{923}{18} & $807$ \down{77} 
\\w/o. PerSam adapt & \dd{817}{35} & \dd{469}{75} & \dd{939}{42} & \dd{911}{56} & $784$ \down{100} \\

\bottomrule
\end{tabular}}

\vspace{0.05in}

\resizebox{0.47\textwidth}{!}{%
\begin{tabular}{lcccc|l}
\toprule

DMControl & Walker  & Cartpole  & Finger & Ball\_in\_cup  & \multirow{2}{*}{\textbf{Average}}\\
(video hard) &Walk & Swingup & Spin & Catch \\
 
\midrule
\textbf{\ours} & \dd{870}{28}& \dd{747}{5}& \dd{932}{22}& \dd{771}{101} & $830$
\\w/o. DINOv2 &   \dd{782}{27}&  \dd{307}{61} & \dd{862}{22}& \dd{404}{97} & $589$  \down{241}
\\w/o. extra points &  \dd{805}{41} & \dd {491}{45}  & \dd{688}{11}& \dd{847}{53} & $708$  \down{122}
\\w/o. refinement & \dd{819}{140} & \dd{717}{75}& \dd{591}{18} &\dd{883}{29} & $753$ \down{77}
\\w/o. PerSam adapt & \dd{803}{26} & \dd{336}{107} & \dd{919}{22} & \dd{802}{112} & $715$ \down{115}\\

\bottomrule
\end{tabular}}
\end{table}

To verify the necessity of designs in \ours, we conduct a series of ablations on 4 DMControl tasks, including two single-object tasks and two multi-object tasks. The quantitative results are in Table~\ref{table:ablations}. We conclude our observations below. More ablations are in the supplementary material.

\noindent\textbf{Usage of DINOv2.} \ours leverages both the DINOv2 encoder and the SAM encoder to extract point features. This design stems from our observation that combining image features from two foundational models enhances the localization capability of point features, as depicted in Figure~\ref{fig:vis feature map}. Our quantitative evaluation results further underscore the efficacy of incorporating DINOv2, particularly in the challenging \textit{video hard} setting. Notably, when we exclude DINOv2, the average performance of \ours drops significantly, decreasing from $830$ to $589$.

\noindent\textbf{Extra points.} In addition to the automatically computed point features, we also include coordinates provided by humans for direct extraction of point features from image features, as detailed in Section~\ref{section: identify}. As illustrated in Table~\ref{table:ablations}, these extra points play a pivotal role in certain tasks, particularly in Cartpole Swingup. Without the inclusion of extra points, the performance of \ours in the \textit{video hard} setting declines from $747$ to $491$. This observation underscores two key insights: (1) the accurate identification of correspondence is of paramount importance, and (2) the \textit{type 1} point features may sometimes fail to provide precise correspondence and our proposed extra points serve as a remedy to mitigate this issue.

\noindent\textbf{Mask refinement.} Similar to extra points, we note the significance of mask refinement particularly in specific tasks such as Finger Spin. While mask refinement generally helps, we have identified a nuanced scenario in the case of Ball\_in\_cup Catch, where mask refinement appears to have a slightly adverse effect. This could be possibly attributed to the characteristics of this task: we observe that the cup frequently moves to the corner of the image, where mask refinement might be even hurtful in identifying such cases.

\noindent\textbf{PerSAM loss.} \ours adopts the efficient parameter fintuning technique from PerSAM~\citep{zhang2023persam} which finetunes 2 learnable weights, for better adaptation to target tasks. We ablate the necessity of the usage. As shown in Table~\ref{table:ablations}, we observe that the usage of this technique is helpful for \ours. From our observation in experiments, this technique helps to find a better understanding of our target object, reducing confusion about what foreground and background are.

\begin{figure}[htbp]
    \centering
    \vspace{-0.1in}
    \includegraphics[width=0.51\textwidth]{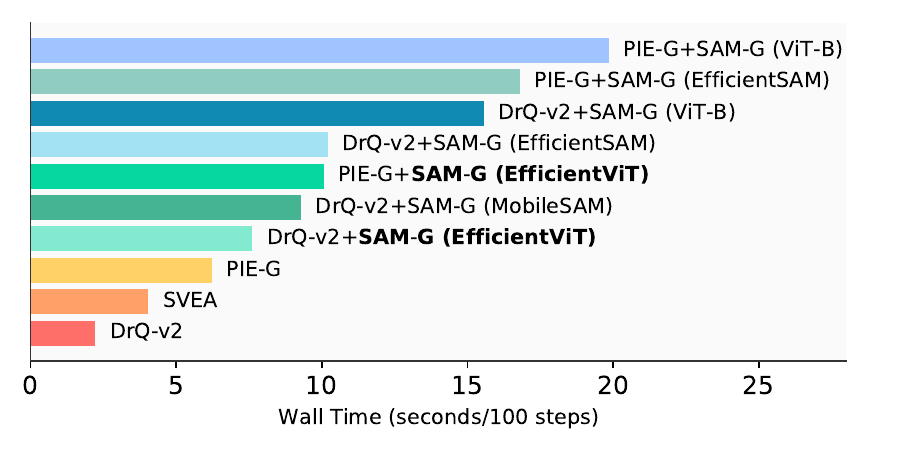}
    \vspace{-0.35in}
    \caption{\textbf{Wall time} of different visual RL algorithms, tested on an NVIDIA A800 GPU on Walker Walk task. The usage of EfficientViT largely improves the inference speed, while it is still costly due to the usage of the large segmentation model.}
    \label{fig:wall time}
    \vspace{-0.1in}
\end{figure}

\noindent\textbf{Speedup inference by EfficientViT.} 
The original ViT model from SAM imposes a large computational overhead when encoding images. To address this issue, we incorporate the EfficientViT~\citep{cai2022efficientvit} architecture. We test the wall time of several visual RL algorithms for comparison, as shown in Figure~\ref{fig:wall time}. It is observed that the adoption of the EfficientViT architecture has made our method more practical, while \ours still exhibits a longer wall time when compared to PIE-G and SVEA. We hope that this overhead can be mitigated in the future through further advancements in model architecture and system optimization.

\noindent\textbf{Different segmentation models.}
\begin{table}[t]
\centering
\caption{\textbf{Ablation on different segmentation models.} We replace the EfficientViT~\citep{cai2022efficientvit} model in \ours with Mask-RCNN~\citep{he2018mask}, SAM (ViT-B)~\citep{kirillov2023sam}, and EfficientSAM~\citep{xiong2023efficientsam} respectively. }
\label{table:ablation segmentation}
\resizebox{0.47\textwidth}{!}{%
\begin{tabular}{lcccc|cc}
\toprule

Ablations & Walker  & Cartpole  & Finger & Ball\_in\_cup  & \multirow{2}{*}{\textbf{Average}}\\
(color hard) &walk & swingup & spin & catch \\
 
\midrule
\textbf{EfficientViT} & \dd{805}{37}& \dd{728}{51}& \dd{739}{32}& \dd{949}{26} & $805$ \\
Mask-RCNN & \dd{566}{124} & \dd{318}{294} & \dd{487}{79} & \dd{448}{163}& $455$ \\
SAM (ViT-B) & \dd{718}{17} & \dd{642}{17} &\dd{752}{44} &\dd{887}{14} &$750$\\
EfficientSAM & \dd{865}{9} & \dd{767}{31} & \dd{896}{17} & \dd{951}{18} & $870$\\
MobileSAM & \dd{906}{22} & \dd{582}{156} & \dd{844}{83} & \dd{617}{128} & 
$737$\\

\bottomrule
\end{tabular}}

\vspace{0.05in}

\resizebox{0.47\textwidth}{!}{%
\begin{tabular}{lcccc|cc}
\toprule

DMControl & Walker  & Cartpole  & Finger & Ball\_in\_cup  & \multirow{2}{*}{\textbf{Average}}\\
(video easy) &Walk & Swingup & Spin & Catch \\
 
\midrule
\textbf{EfficientViT} & \dd{887}{33}& \dd{855}{6}& \dd{937}{21}& \dd{858}{79} & $884$ \\
Mask-RCNN & \dd{650}{161} & \dd{257}{60} & \dd{482}{29} & \dd{536}{48} & $481$\\
SAM (ViT-B) & \dd{820}{27} & \dd{624}{8} & \dd{940}{34} &\dd{966}{5} &$838$\\
EfficientSAM & \dd{869}{25} & \dd{824}{10} & \dd{959}{18} & \dd{951}{16} & $901$\\
MobileSAM & \dd{883}{67} & \dd{650}{105} & \dd{932}{30} & \dd{790}{81} & 
$814$\\

\bottomrule
\end{tabular}}

\vspace{0.05in}

\resizebox{0.47\textwidth}{!}{%
\begin{tabular}{lcccc|cc}
\toprule

DMControl & Walker  & Cartpole  & Finger & Ball\_in\_cup  & \multirow{2}{*}{\textbf{Average}}\\
(video hard) &Walk & Swingup & Spin & Catch \\
 
\midrule
\textbf{EfficientViT} & \dd{870}{28}& \dd{747}{5}& \dd{932}{22}& \dd{771}{101} & $830$\\
Mask-RCNN & \dd{62}{20} & \dd{150}{30} & \dd{37}{5} & \dd{100}{41} & $87$\\
SAM (ViT-B) & \dd{735}{10} & \dd{517}{21} & \dd{896}{35} &\dd{927}{2}&$769$\\
EfficientSAM & \dd{872}{5} & \dd{732}{33} & \dd{862}{15} & \dd{901}{35} & $842$\\
MobileSAM & \dd{867}{57} & \dd{543}{95} & \dd{858}{24} & \dd{534}{84} & 
$701$\\
\bottomrule
\end{tabular}}
\vspace{-0.1in}
\end{table}

One key design of \ours is to apply SAM, the most powerful segmentation foundation model as far as we know. We have also tried other segmentation models for comparison, including a Mask R-CNN~\citep{he2018mask} pre-trained on COCO, a ViT-B version of original SAM, EfficientSAM~\citep{xiong2023efficientsam}, and MobileSAM~\citep{zhang2023mobileSAM}. EfficientSAM and MobileSAM are two distilled versions of SAM for faster inference. We use the EfficientSAM-S model.

\begin{figure}[t]
    \includegraphics[width=\linewidth]{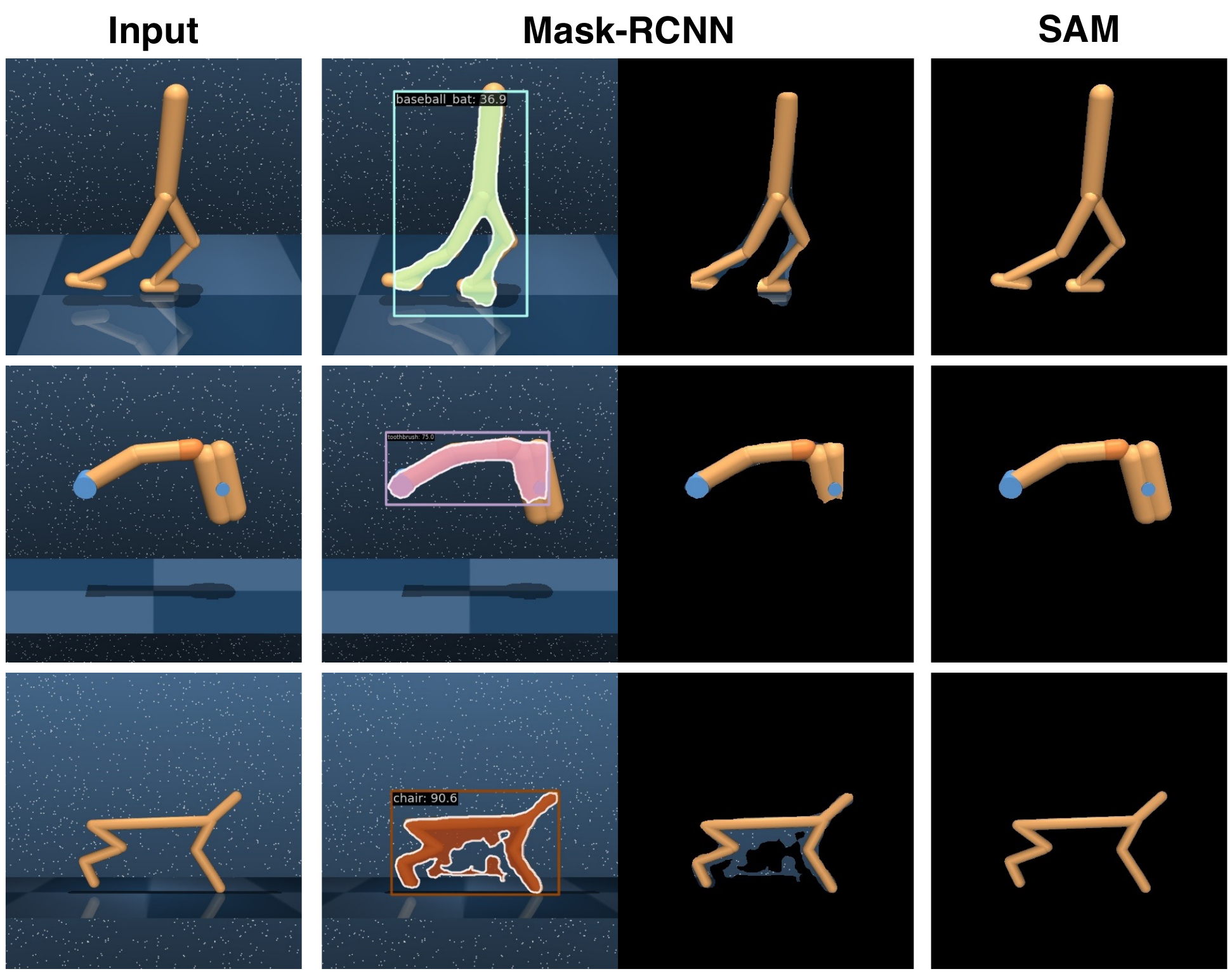}
  \vspace{-0.3in}
  \caption{\textbf{Visualization of segmentation from Mask-RCNN and SAM for comparison.} Mask-RCNN only roughly detects the object in the center, while SAM produces high-quality fine-grained masks.}
  \label{fig:maskrcnn}
  \vspace{-0.1in}
\end{figure}

As shown in Table~\ref{table:ablation segmentation}, \ours that leverages an Efficient-ViT version of SAM largely surpasses that with Mask-RCNN. This is not surprising since Mask-RCNN can not make competitive segmentation results compared to SAM, as visualized in Figure~\ref{fig:maskrcnn}.  We also observe that the original SAM (ViT-B) works slightly worse than \ours, since \ours is using the Efficient ViT-L1 model, which is distilled from the more accurate ViT-H SAM model. EfficientSAM is also trained to distill the ViT-H model in SAM and it gets an even better result than \ours, while due to its larger wall time (see Figure~\ref{fig:wall time}), we apply EfficientViT in \ours. MobileSAM achieves similar wall time but is less accurate, compared to EfficientViT.

\section{Conclusion}


In this work, we introduce \oursfull, a framework that harnesses vision foundation models to find correspondence and subsequently leverages Segment Anything Model (SAM) to segment out task-relevant objects for the benefit of visual reinforcement learning (RL) agents. The high-quality masked images produced by SAM are directly utilized by agents. We conduct evaluations of \ours on the generalization benchmark of 11 visual RL tasks and observe the robust generalization capabilities of \ours across settings of varying difficulty. Notably, in the most challenging setting, \textit{video hard}, we achieve relative improvements of $44\%$ and $29\%$ across two domains when compared to the state-of-the-art method. Our work highlights the significance of leveraging vision foundation models for enhancing generalization in decision-making.

One limitation of our work is the extended wall time when employing SAM. To mitigate this issue, we have incorporated EfficientViT in this work, and we anticipate that future advancements in machine learning systems and architectures will further address this challenge.

\section*{Acknowledgement}

This work is supported by National Key R\&D Program of China (2022ZD0161700).

\clearpage

{
    \small
    \bibliographystyle{ieeenat_fullname}
    \bibliography{main}
}

\clearpage
\setcounter{page}{1}
\maketitlesupplementary
\appendix

\section{Implementation Details}
In this section, we describe more implementation details of \ours, mainly describing how to obtain embeddings, point features, and SAM prompts. An overview of \ours has been shown in Figure~\ref{fig:overview}. Our official implementation is available at \url{https://github.com/wadiuvatzy/SAM-G}.

\noindent\textbf{Obtain feature embeddings.} To get the point prompts, we first extract feature embeddings from the visual encoder of SAM and DINOv2, and the resulting embedding is used to compute the similarity map. Taking the image observation of size $84\times 84 \times 3$ as input, the encoder of SAM resizes the image into $1024\times 1024\times 3$ and obtains an embedding of size $64 \times 64 \times 256$. Similarly, the encoder of DINOv2 resizes the image into $448\times 448\times 3$ and obtains an embedding of size  $32\times 32 \times 768$. The DINOv2 embedding is resized to $64\times 64\times 768$ for alignment.



\noindent\textbf{Obtain point features.} Given a pair of image and its mask from the training environment, we obtain 2 types of target point feature as shown in Figure~\ref{fig:fetch point feature}. \textit{Type 1} point features are computed on the masked embeddings using spatial average and max operations. \textit{Type 2} point features are fetched directly from original embeddings coordinated by human-given points. We observe that \textit{type 2} point features are more necessary for multi-object scenarios and confusing backgrounds, and for easier settings, \textit{type 2} point features are shown to be not very necessary.

\noindent\textbf{Prompt for segmentation.} Once we prepare the point features, we are ready to segment images. Take the case with 1 \textit{type 1} point feature and 3 \textit{type 2} point features as an example. For an image we want to segment, we first get the feature embeddings through the method discussed above. And then we will use \textit{mask decoder} of SAM 3 times to get the final segmentation result.

\noindent\textbf{\textbullet\ 1st segmentation.} for each point feature, calculating the normalized inner product with the feature embedding gives us two $64 \times 64$ similarity maps (one from SAM feature and the other from DINOv2 feature). By taking the mean of each pair of two similarity maps, we get 4 final similarity maps: one from \textit{type 1} point feature and three from \textit{type 2} point feature. All the 4 similarity maps give us a positive point prompt by finding the maxima in each map, while only the similarity map from \textit{type 1} point feature gets the minima as a negative point prompt. We do not use \textit{type 2} similarity map to get negative point prompts because \textit{type 2} point feature only contains local information, the minima may lie on another object we want. Feeding those 5 point prompts together with the feature embedding got before to \textit{mask decoder} of SAM is the final step to get the first result.

\noindent\textbf{\textbullet\ 2nd and 3rd segmentation}. After the first segmentation, we can get 3 masks (multimask return by SAM) together with their mask logits. Conducting a weighted summation of those mask logits with our trained 3 weights gives us a rough mask whose bounding box is also easily got. We can find the minima of \textit{type 1} similarity map inside the box and take that as another negative point prompt (may be the same point as the previous negative prompt). Further we will feed those 6 point prompts together with the box, the logits into \textit{mask decoder} of SAM to get the 2nd result. For the last time of segmentation, repeating similar process in 2nd segmentation, we can get a bounding box of a rough mask, a mask logits and another negative prompt. Once again, feeding 7 point prompt, 1 box and 1 mask logits to the \textit{mask decoder} of SAM gives out 3 masks. We choose the one with highest score (given by SAM) as our final segmentation result.

\section{Descriptions of Tasks}
\label{appendix: task and generalization description}

Our generalization setting follows \citep{hansen2021soda,yuan2023rlvigen}, where the \textit{color} setting changes the background, object color and table texture, while the \textit{video} setting changes the background to a natural video and introduces moving light. Compared with \textit{color easy} setting, \textit{color hard} setting presents a higher level of randomness and complexity, with more varied and unpredictable visual features. Similarly, \textit{video easy} setting has simpler video background dynamics, while \textit{video hard} setting consists of intricate and swiftly alternating video backgrounds which are substantially dissimilar to the training setting.

We describe the tasks used in this work as follows, mainly from DMControl~\citep{tassa2018dmc} and Adroit~\citep{rajeswaran2017dapg}.


\begin{itemize}
\item  \textit{Walker, walk} ($\mathbf{a}\in\mathbb{R}^{6}$). A planar walker that is rewarded for walking forward successfully at a set speed. Dense rewards.
\item  \textit{Walker, stand} ($\mathbf{a}\in\mathbb{R}^{6}$). A planar walker that is rewarded for  maintaining a vertical position and a steady height above a specified minimum. Dense rewards.
\item  \textit{Cartpole, swingup}($\mathbf{a}\in\mathbb{R}$). Swing up and stabilize a free-standing rod by exerting forces on a cart at the foundation. The agent is rewarded for keeping the rod aligned within a set angular limit. Dense rewards. Dense rewards.
\item \textit{Cheetah, run} ($\mathbf{a}\in\mathbb{R}^{6}$). A planar cheetah model that is rewarded for sprinting forward rapidly, with the objective of achieving and maintaining a high velocity. Dense rewards.
\item \textit{Hopper, stand} ($\mathbf{a}\in\mathbb{R}^{4}$). A one-legged planar hopper that is rewarded for achieving and maintaining an upright position, with its torso held vertically to a minimal height. Dense rewards.
\item \textit{Finger, spin} ($\mathbf{a}\in\mathbb{R}^{2}$). A planar robotic finger that is rewarded for spinning a body affixed to a surface, with the objective of achieving and maintaining a continuous rotational velocity. Dense rewards.
\item \textit{Ball\_in\_cup, catch} ($\mathbf{a}\in\mathbb{R}^{2}$). A motorized planar container that is rewarded for oscillating and catching a ball tethered to its base by a string. Sparse rewards.
\item \textit{Quadruped, walk} ($\mathbf{a}\in\mathbb{R}^{12}$). A four-legged robotic entity that is rewarded for ambulating forward, aiming to achieve a targeted gait and speed. Dense rewards.
\item \textit{Door} 
($\mathbf{a} \in\mathbb{R}^{28}$). The task to be completed consists on unlocking the door and swing the door. The task is considered complete when the door touches the door stopper. Sparse rewards.
\item \textit{Pen}
($\mathbf{a} \in\mathbb{R}^{24}$). The task to be completed consists on repositioning the blue pen to match the orientation at the green target pen which is randomly chosen from all configurations. The task is considered complete when the orientations match with some tolerance. Sparse rewards.
\item \textit{Hammer}
($\mathbf{a} \in\mathbb{R}^{26}$). The task to be completed consists on picking up a hammer and drive a randomly positioned nail into a board. The task is considered complete when the neil is entirely in the board. Sparse rewards.
\end{itemize}


\section{Hyperparameters}
\label{appendix: hyperparam}
Our algorithms are mainly based on DrQ-v2 and PIE-G and most of the training hyperparameters are the same. Table~\ref{table:hyperparam} shows detailed hyperparameters. Task-relevant parameters are in Table~\ref{table:task parameters}.


\begin{table}[t]
\centering
\caption{\textbf{Hyperparameters} for \ours.}
\label{table:hyperparam}
\vspace{-0.1in}
\resizebox{0.5\textwidth}{!}{%
\begin{tabular}{lcccccc}
\toprule

Hyperparameter &   DMControl-GB & Adroit \\
 
\midrule
Input size & 84 $\times$ 84 & 84 $\times$ 84 (door, hammer), 160$\times$160 (pen)\\
Discount factor $\gamma$ & 0.99 & 0.99\\
Action repeat & 2 & 2\\
Frame stack & 3 &  3\\
Learning rate &  1e-3 & 1e-4\\
Random shifting padding & 4 & 4\\
Episode length & 1000 & 200 (door, hammer), 100 (pen)\\
Evaluation episodes & 100 & 100\\
Batch size & 128 & 256 \\
Replay buffer size & 5e5 & 1e6 \\
Optimizer &   Adam & Adam \\

\bottomrule
\end{tabular}}
\end{table}

\begin{table}[t]
\centering
\caption{\textbf{Task parameters} for DMControl and Adroit tasks. }
\label{table:task parameters}
\vspace{-0.1in}
\resizebox{0.5\textwidth}{!}{%
\begin{tabular}{lcccccc}
\toprule

Task & Action Dim & Action Repeat & \# Objects & \# Extra points  & \# Frames \\
 
\midrule
Walker Walk & 6  &2 & 1 & 3 & 1M\\
Walker Stand & 6 & 2& 1 & 3 & 1M\\
Cartpole Swingup & 1 &2 & 1 & 3 & 1M  \\
Ball\_in\_cup Catch & 2 & 2 & 2 & 6 & 1M\\
Hopper Stand & 4 & 2& 1 & 3 & 1M \\
Finger Spin & 2 & 2 & 1 & 3 & 1M\\
Cheetah Run & 6 & 2& 1 & 3 & 3M\\
Quadruped Walk & 12 & 2 & 1 & 3 & 3M \\
\midrule
Door & 28 & 2 & 2 & 6 & 1M \\
Hammer & 26 & 2 & 3 & 9 & 1M \\
Pen & 24 & 2 & 2 & 6 & 2M \\

\bottomrule
\end{tabular}}
\end{table}

\section{More Ablations}
\label{appendix: more ablations}

\subsection{Higher Resolution for Pen}
\begin{figure}[htbp]
    \centering
    \vspace{-0.1in}
    \includegraphics[width=0.4\textwidth]{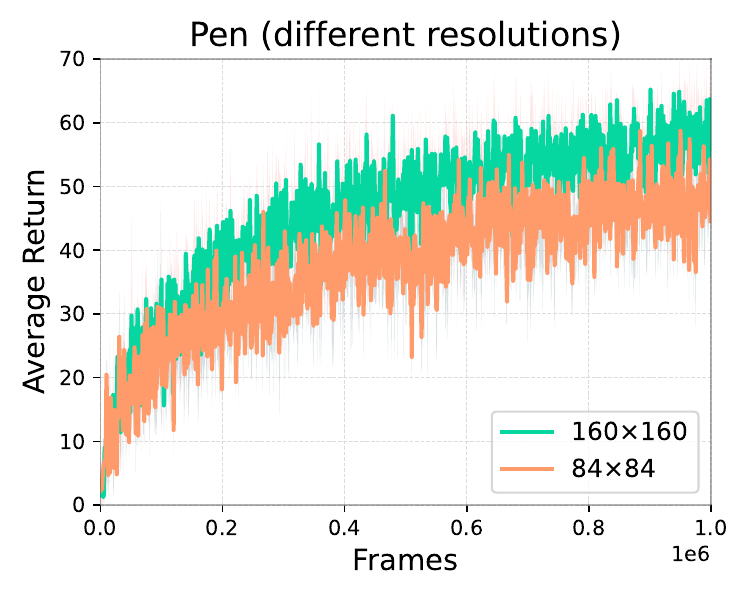}
    \vspace{-0.2in}
    \caption{\textbf{Training curves in different resolution in Pen.} Mean of 3 random seeds, shaded area is $\pm1$ std. Training in $160\times 160$ input get better performance than training in $84\times84$ for \ours.}
    \label{fig:pen high resolution}
    \vspace{-0.1in}
\end{figure}
\label{appendix: discussion about pen}

\begin{figure}[htbp]
  \begin{subfigure}{0.5\linewidth} 
    \includegraphics[width=\linewidth]{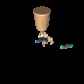} 
    \caption{$84\times 84$}
    \label{fig:sub1}
  \end{subfigure}%
  \begin{subfigure}{0.5\linewidth} 
    \includegraphics[width=\linewidth]{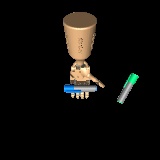} 
    \caption{$160\times 160$}
    \label{fig:sub2}
  \end{subfigure}
  \vspace{-0.3in}
  \caption{\textbf{Visualization of segmentation under different resolutions.} SAM segments more accurately under a higher resolution.}
  \label{fig:different resolution}
  \vspace{-0.2in}
\end{figure}


\begin{table}[htbp]
\centering
\caption{\textbf{The low resolution result for pen.} All the resolution of the visual observation are $84\times 84$.}
\label{table:pen low resolution}
\resizebox{0.47\textwidth}{!}{%
\begin{tabular}{lcccc|cc}
\toprule

Pen  & Color & Color  & Video & Video & \multirow{2}{*}{\textbf{Average}}\\

high resolution & easy & hard & easy & hard \\
 
\midrule
VRL3 & \dd{0.1}{0.0} & \dd{0.1}{0.0} & \dd{0.0}{0.0} & \dd{3.6}{0.9} & $1.0$\\
SVEA  & \dd{55.0}{8.6} & \dd{47.0}{6.8} & \dd{50.2}{8.6} & \dd{46.8}{9.7} & $49.8$\\
SGQN  & \dd{51.4}{18.1} & \dd{36.8}{13.7} & \dd{51.4}{18.1} & \dd{54.0}{3.7} & $48.4$\\
PIE-G  & \dd{70.6}{2.9} & \dd{57.4}{4.2} & \dd{67.0}{8.8} & \dd{56.6}{6.3} & $62.9$\\
\textbf{\ours}  & \dd{60.0}{12.8} & \dd{53.0}{5.0} & \dd{61.3}{9.3} & \dd{55.3}{4.2} & $57.4$\\

\bottomrule
\end{tabular}}
\end{table}

In Table~\ref{table:adroit}, we mentioned that \ours falls short of performance in the Pen task in $84\times 84$ resolution. The detailed evalation result can be found in Table~\ref{table:pen low resolution}.
 We postulate that the unsatisfying performance of \ours in the Pen task may be attributed to the low image resolution. In the Pen task, the dexterous hand must execute precise manipulation, and the challenges in segmenting objects become pronounced in low-resolution images. To substantiate our hypothesis and enhance \ours' performance, we increase the input image resolution from $84\times 84$ to $160\times 160$ for improved segmentation. The training progress is shown in Figure~\ref{fig:pen high resolution}, and the final corresponding evaluation results are presented in Table~\ref{table:adroit}. We also compare the segmentation results in Figure~\ref{fig:different resolution}.

Notably, the success rate in the \textit{train} setting has substantially improved, rising from $69\%$ in the low-resolution setting to an impressive $82\%$ in the high-resolution setting. Moreover, with the adoption of this higher resolution, \ours has now significantly outperformed all other baselines, providing strong validation for our initial hypothesis.

\end{document}